\documentclass[10pt,conference]{IEEEtran} 
\IEEEoverridecommandlockouts
\usepackage{cite}
\usepackage{xspace} 
\usepackage{amsmath,amssymb,amsfonts}
\usepackage{algorithmic}
\usepackage{graphicx}
\usepackage{textcomp}
\usepackage{xcolor}
\usepackage{stfloats}
\usepackage{balance}
\usepackage{hyperref}
\usepackage{soul}
\usepackage{enumitem}
\usepackage{booktabs}
\usepackage{caption}
\usepackage{siunitx}
\usepackage{tabularx}
\usepackage{subcaption}
\usepackage{tcolorbox}
\usepackage{array}
\usepackage[retain-zero-uncertainty=true]{siunitx}
\newcommand\approach{DILLEMA\xspace}

\def\BibTeX{{\rm B\kern-.05em{\sc i\kern-.025em b}\kern-.08em
    T\kern-.1667em\lower.7ex\hbox{E}\kern-.125emX}}

\begin{document}

\title{
DILLEMA: Diffusion and Large Language Models for Multi-Modal Augmentation\\
\thanks{This research was partially supported by project EMELIOT, funded by MUR under the PRIN 2020 program (Contract 2020W3A5FY).
}
}
%

\author{
\IEEEauthorblockN{
  Luciano Baresi,
  Davide Yi Xian Hu,
  Muhammad Irfan Mas'udi,
  Giovanni Quattrocchi
}
\IEEEauthorblockA{
  \textit{Dipartimento di Elettronica, Informazione e Bioingegneria, Politecnico di Milano},
  Milan, Italy \\
  luciano.baresi@polimi.it, davideyi.hu@polimi.it, muhammadirfan.masudi@mail.polimi.it, giovanni.quattrocchi@polimi.it
}
}

\maketitle

\thispagestyle{plain}
\pagestyle{plain}

\IEEEpeerreviewmaketitle 

\begin{abstract}
Ensuring the robustness of deep learning models requires comprehensive and diverse testing. Existing approaches, often based on simple data augmentation techniques or generative adversarial networks, are limited in producing realistic and varied test cases. To address these limitations, we present a novel framework for testing vision neural networks that leverages Large Language Models and control-conditioned Diffusion Models to generate synthetic, high-fidelity test cases. Our approach begins by translating images into detailed textual descriptions using a captioning model, allowing the language model to identify modifiable aspects of the image and generate counterfactual descriptions. These descriptions are then used to produce new test images through a text-to-image diffusion process that preserves spatial consistency and maintains the critical elements of the scene. We demonstrate the effectiveness of our method using two datasets: ImageNet1K for image classification and SHIFT for semantic segmentation in autonomous driving. The results show that our approach can generate significant test cases that reveal weaknesses and improve the robustness of the model through targeted retraining. We conducted a human assessment using Mechanical Turk to validate the generated images. The responses from the participants confirmed, with high agreement among the voters, that our approach produces valid and realistic images.
\end{abstract}


\section{Introduction}
\label{sec:intro}
Testing deep learning-based systems (DL)~\cite{GoodBengCour16} is a complex and critical task that shares similarities with traditional software testing, but presents unique challenges due to the data-driven nature of these systems.

These systems operate in high-dimensional input spaces, such as pixel values for images or token sequences for text. The large size and complexity of this input space make it practically impossible to test all possible inputs. Traditional testing techniques cannot cover such large input spaces, and identifying corner cases that could cause the model to fail requires specialized methods.
In addition, determining the correct output is not always straightforward, especially when dealing with complex tasks such as image classification or autonomous driving. The lack of a clear oracle, also known as the \textit{Oracle Problem}, makes it difficult to determine whether the system behavior is correct, as there is often no ground truth for comparison.
Furthermore, DL models are typically made up of many layers of interconnected neurons, making them complex and difficult to interpret~\cite{DBLP:journals/tse/ZhangHML22}.
As a consequence, they are often treated as black boxes that learn complex representations through data.

Recently, a great deal of effort has been dedicated to using metamorphic testing~\cite{DBLP:conf/icse/ZhouLKGZ0Z020, DBLP:journals/pacmse/ChenJYGZC24} to address the aforementioned challenges.
Metamorphic testing evaluates the behavior of a DL model by systematically applying transformations to input data and examining the corresponding output changes. This approach relies on metamorphic relationships that formally define how input modifications affect the output~\cite{DBLP:conf/icse/DingKH17}.
Metamorphic testing for vision neural networks addresses the oracle problem by leveraging metamorphic relations to generate new test cases. This approach applies systematic transformations to existing test data while preserving ground truth labels (i.e., without affecting the expected output), enabling comprehensive testing without an explicit oracle.

Metamorphic testing has recently been employed in several approaches. Tian et al.~\cite{DBLP:conf/icse/TianPJR18} used transformations (such as adjustments to brightness, rotation, and blurring) on existing images to generate new test cases for DL-based autonomous driving systems. Although these transformations capture different behaviors of camera sensors, they do not represent realistic variations of the surrounding environment.
Zhang et al.~\cite{DBLP:conf/kbse/ZhangZZ0K18} applied Generative Adversarial Networks (GANs) to validate the behavior of the model in diverse scenarios. Although GANs provide an effective and scalable approach for generating large numbers of diverse test cases, they require a dedicated dataset for each target scenario and an ad-hoc training process to teach the generative model to map images from one domain to another (e.g., transforming images with sunny weather into images with rainy weather). This makes GAN-based testing resource-intensive, as it requires significant manual effort to create and synthesize new domains.

This paper introduces \approach (\textbf{DI}ffusion model and \textbf{L}arge \textbf{L}anguag\textbf{E} \textbf{M}odel for \textbf{A}ugmentation), a framework designed to enhance the robustness of DL applications by automatically augmenting existing image datasets. \approach utilizes a \textit{Captioning Model} (CM) to generate textual descriptions from input images. It uses a \textit{ Large Language Model} (LLM) to generate new descriptions, and a controllable \textit{Diffusion Model} (DM) to generate realistic high-quality images. Specifically, by taking advantage of pre-trained models that have been trained on large amounts of data, \approach generalizes well across various scenarios and datasets without the need for ad-hoc training (as required by approaches based on GANs~\cite{DBLP:conf/kbse/ZhangZZ0K18}).
We evaluated \approach using two popular datasets, ImageNet1K~\cite{DBLP:conf/cvpr/DengDSLL009} and SHIFT~\cite{DBLP:conf/cvpr/SunSPWGSTY22}.
The evaluation of DILLEMA covered multiple aspects, including the validity and hallucination rates of the generative models used. For example, the results show that the generated test cases maintained high validity, with more than $99.7\%$ augmented ImageNet1K images that preserved their original labels according to human evaluators.
Furthermore, empirical results demonstrate that the generated test suites uncover significantly more vulnerabilities compared to existing datasets. \approach revealed error rates more than $15$ times higher on ImageNet1K than the existing original test set. Furthermore, retraining with the augmented test cases improved robustness by up to $52.27\%$. 

The main contributions of this paper are as follows.

\noindent\textbf{Novel Metamorphic Testing Pipeline}.
    A framework to enhance DL models by automatically augmenting image datasets and generating new images using a combination of Captioning Model, Large Language Model, and Diffusion Model.
    
\noindent\textbf{Testing Dataset}.
    We released two additional datasets for testing DL applications: $125,000$ test cases for ImageNet1K classification and $10,000$ for SHIFT autonomous driving.
    
\noindent\textbf{Comprehensive Evaluation}.
    We assessed the approach's effectiveness and the realism of its generated images.

The remainder of this paper is structured as follows, \autoref{sec:solution} presents DILLEMA, \autoref{sec:eval} shows the empirical evaluation, \autoref{sec:related} introduces related work, and \autoref{sec:conclusion} concludes the paper.

\section{Methodology}
\label{sec:solution}

This paper presents \approach, a framework that improves the robustness of DL-based systems by generating realistic test images from existing datasets.
\approach leverages recent advances in text and visual models~\cite{DBLP:conf/cvpr/RombachBLEO22} to generate accurate synthetic images to test DL-based systems in scenarios that are not represented in the existing testing suite.

\begin{figure*}
    \centering
    \includegraphics[width=\linewidth]{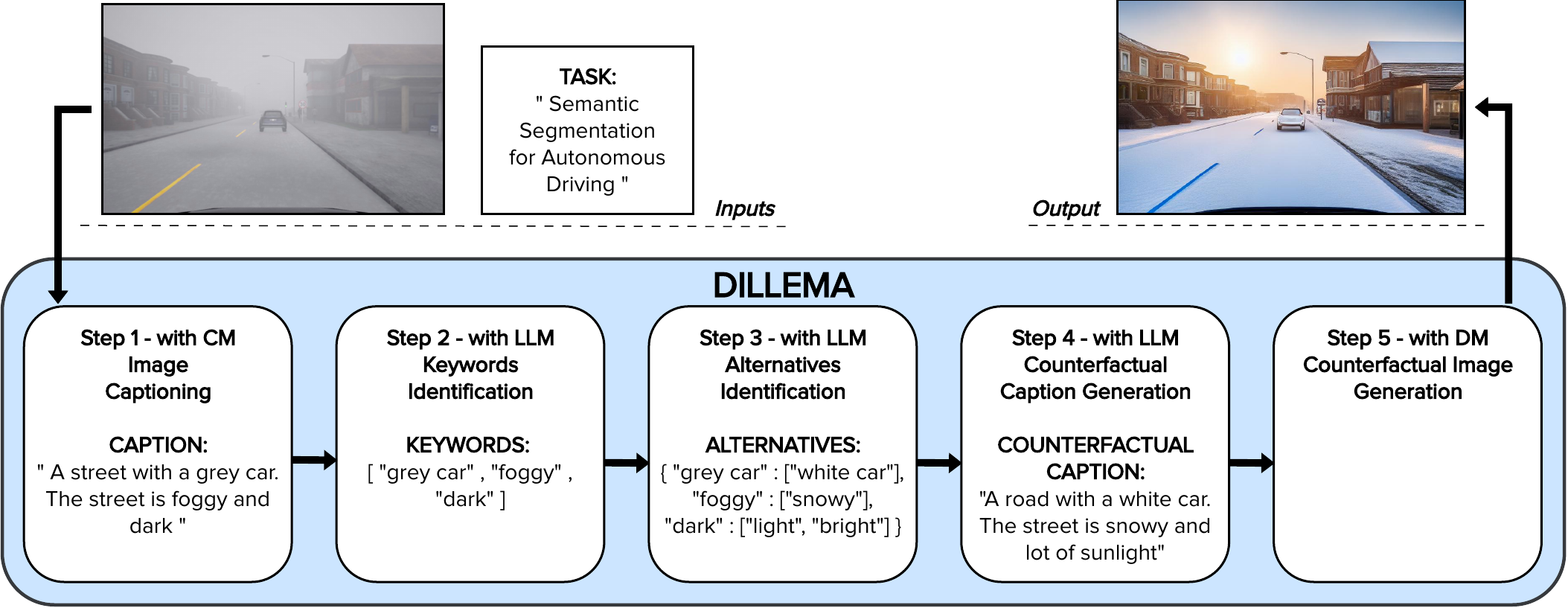}
    \caption{\approach.}
    \label{fig:dillema}
    \vspace{-6mm}
\end{figure*}

The proposed methodology, as shown in \autoref{fig:dillema}, consists of five steps.
The input of our approach is an image (from the existing test cases) along with a textual description of the task assigned to the DL-based system. The output is a modified version of the input image based on new conditions. 

\subsection{Image Captioning}

The first step of \approach involves image captioning, which is the process of converting a given image to its textual description. The objective is to enable the application of recent advances in natural language processing to images. To achieve this, \approach brings the images into the textual domain, where language models can operate effectively.

Captions are generated as multi-sentence descriptions to capture key elements and provide a detailed representation of the image. Each sentence focuses on a different aspect of the scene, capturing a range of elements such as objects, environments, and contextual relationships. This approach increases the likelihood of capturing important details that a single-sentence description might miss, providing a more comprehensive textual description for the subsequent steps.

\subsection{Keyword Identification}

Once the image is converted into textual descriptions through the captioning process, the next step in \approach is Keyword Identification. This step aims to identify which elements of the image can be safely modified without altering the overall meaning or the primary task (e.g., object classification, semantic segmentation) associated with the image. 

\begin{figure}[]
\centering
    \begin{subfigure}{.49\linewidth}
    \centering
    \includegraphics[width=\textwidth]{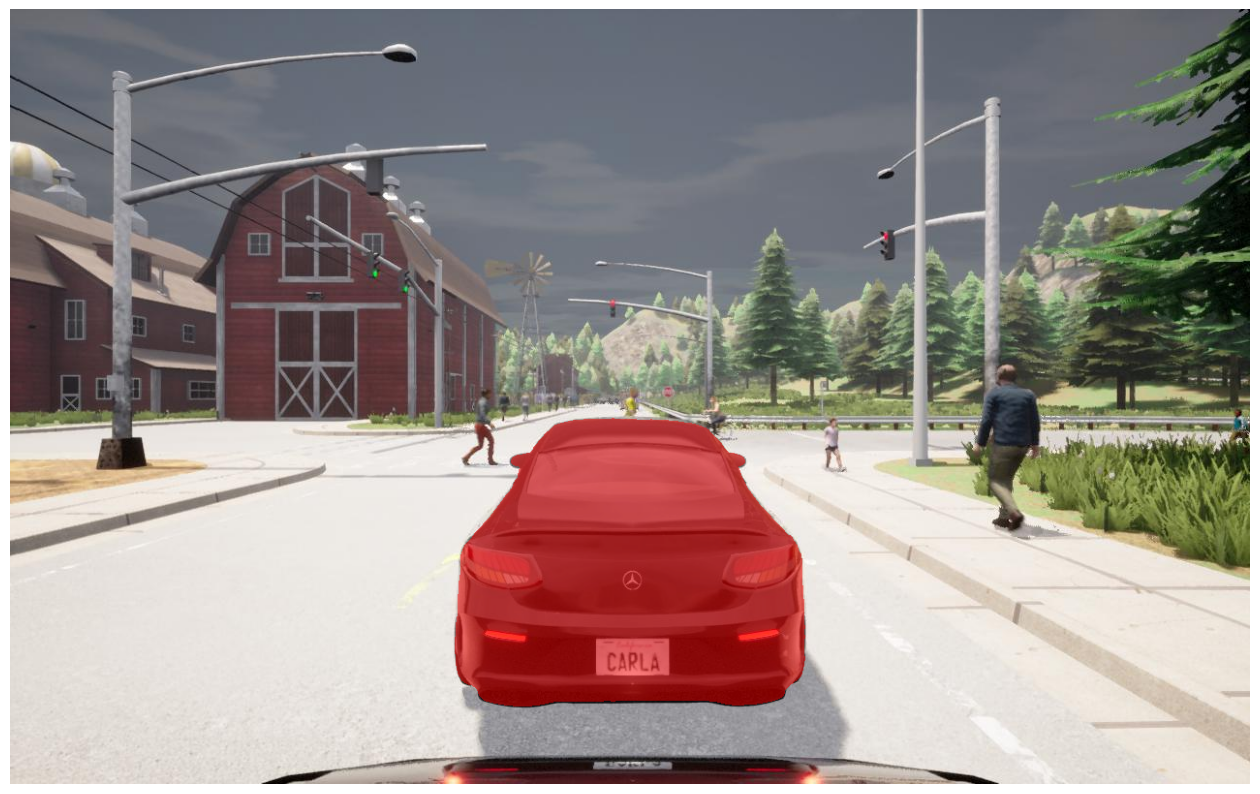}
    \caption{Classification Task.}
    \label{fig:classification}
    \end{subfigure}
    \begin{subfigure}{.49\linewidth}
    \centering
    \includegraphics[width=\textwidth]{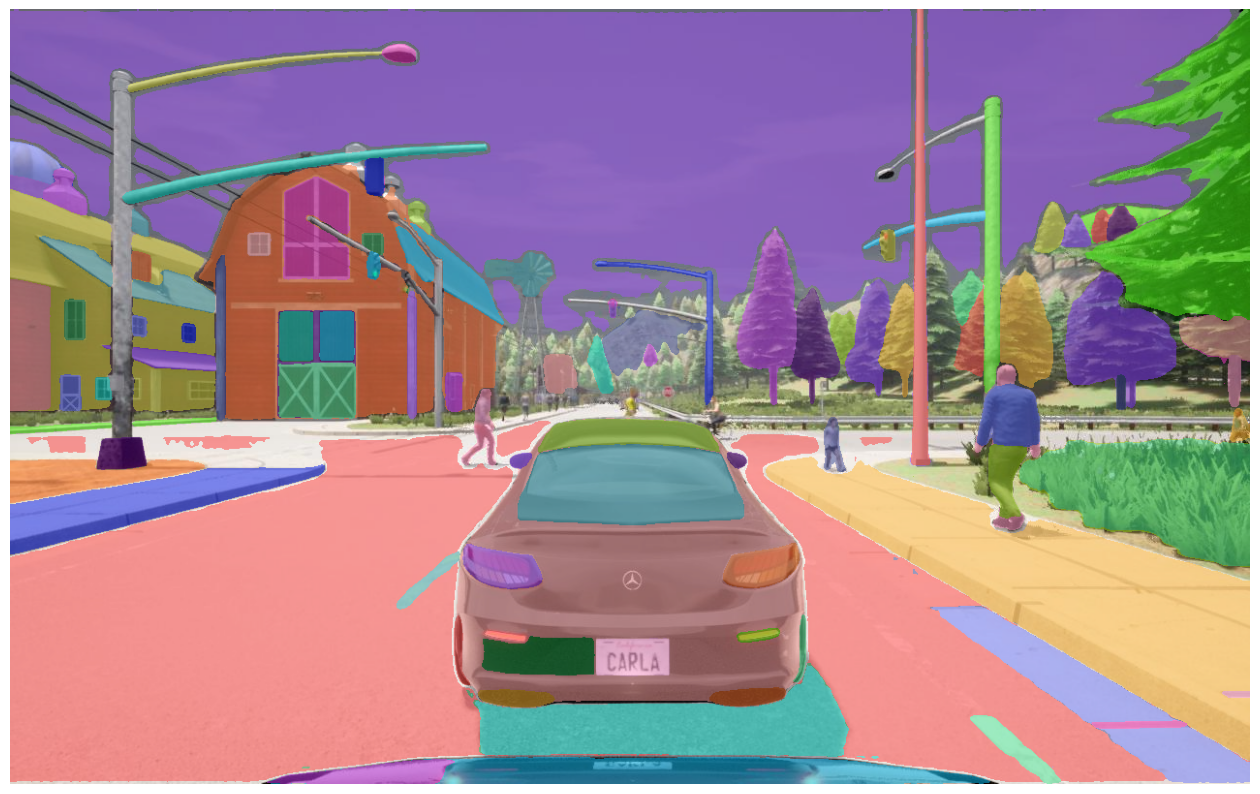}
    \caption{Segmentation Task.}
    \label{fig:semantic_segmentation}
    \end{subfigure}
\caption{
Label Preservation in Autonomous Driving Tasks.
}
\label{fig:classification_and_segmentation}
\vspace{-6mm}
\end{figure}

In this phase, the LLM is used to analyze the captions generated in the previous step and identify a set of keywords that can potentially be altered. These keywords represent modifiable aspects of the image, such as colors, weather conditions, or object properties, while excluding core elements that are essential to the task. For example, when dealing with an image classification task involving a ``car'', altering the background color or lighting usually does not modify the label. Conversely, in a semantic segmentation task focused on road scenes, the road and critical objects (cars, pedestrians, traffic signals) must remain present, though certain attributes (e.g., color, weather conditions) can still be changed. By defining the task explicitly in the prompt, we ensure that only permissible alterations are suggested by the LLM.
\autoref{fig:classification_and_segmentation} illustrates how the constraints differ between classification (\autoref{fig:classification}) and segmentation (\autoref{fig:semantic_segmentation}). In classification, the focus is on identifying and preserving the labeled object (\emph{car}), while in segmentation, multiple objects must remain for valid ground-truth labels.

The process of identifying these keywords is guided by task constraints. \approach prompts the LLM with a specific task-related query, such as:

\begin{tcolorbox}[arc=.3em,left=.3em,right=.3em,top=.3em,bottom=.3em]
\begin{center}
\begin{minipage}[t]{.99\linewidth}
\textbf{Prompt}: \textit{
Given the task $<$TASK$>$ and an image described by the caption $<$CAPTION$>$, what are the key elements that can be modified in the caption so that the ground truth corresponding to the image does not change?
}
\end{minipage}
\end{center}
\end{tcolorbox}

Note that this represents an example of the prompts used in \approach, intended to clarify the type of information that we request from the LLM. To improve the effectiveness of the prompt, various advanced strategies can be adopted. For example, as detailed in \autoref{sec:eval:setup}, we configured \approach to use a one-shot in-context learning prompting strategy, allowing the LLM to provide better results by including an example within the prompt.

The identification of keywords is designed to be flexible and adaptable for different tasks. The LLM relies on its internal knowledge to evaluate the contextual relevance of each word in the caption, taking into account both syntactic and semantic relationships. For example, if the task is semantic segmentation in an autonomous driving scenario, elements such as road conditions, lighting, or vehicle color may be identified as modifiable keywords, while objects essential to the task, such as vehicles themselves, remain unchanged.

\subsection{Alternative Identification}

In this phase, the LLM is leveraged to generate alternatives for the identified keywords, providing variations that can be applied to the image without altering the overall task.

The goal of this step is to explore different possibilities for modifying the elements flagged in the previous step, such as changing the color of objects, adjusting environmental conditions (e.g., weather), or altering minor details, while keeping the core structure and purpose of the image intact. For example, if the keyword ``foggy'' was identified as a modifiable attribute in the caption ``a car driving down a foggy street'', the LLM could suggest alternatives like ``rainy'' or ``snowy''.
To execute this, \approach generates a prompt asking the LLM to propose alternatives for the identified keywords. 

The main challenge in this phase is to introduce meaningful variations to the image while keeping its semantic meaning intact. The LLM plays a key role by generating alternatives that align with the original caption and task, avoiding changes that could shift the focus of the task. We take advantage of the ability of the LLM to understand contextual subtleties to avoid proposing changes to critical elements such as replacing ``car'' with ``bicycle'' in a vehicle detection scenario. An example of a prompt used in this phase is:

\begin{tcolorbox}[arc=.3em,left=.3em,right=.3em,top=.3em,bottom=.3em]
\begin{center}
\begin{minipage}[t]{.99\linewidth}
\textbf{Prompt}: \textit{
Given the task $<$TASK$>$ and an image described by the caption $<$CAPTION$>$, what are the possible alternatives for these keywords $<$KEYWORDS$>$?
}
\end{minipage}
\end{center}
\end{tcolorbox}

This process focuses on generating contextually relevant and diverse modifications, allowing the system to produce meaningful test cases for the DL model at hand. The alternatives proposed for each keyword enable \approach to explore different conditions or attributes of objects, broadening the range of scenarios included in the original dataset.

\subsection{Counterfactual Caption Generation}

This phase is responsible for creating new textual descriptions, or counterfactual captions, by applying the alternatives generated in the previous step. These counterfactual captions describe how the image would look if certain elements were modified, enabling the system to explore new scenarios while preserving the core context of the original image.

In this step, the LLM takes the original caption and replaces the identified keywords with the newly generated alternatives. The goal is to produce a new version of the caption that reflects the desired modifications without changing the essential meaning of the image. For example, if the original caption was ``a gray car driving down a foggy street'', and the alternatives generated for the keywords ``gray car'' and ``foggy'' were ``red car'' and ``snowy'', the new counterfactual caption would be ``A red car driving down a snowy street''.

The amount of edits in the new prompt can be controlled by limiting the number of alternatives applied when generating the counterfactual captions. For example, applying only one alternative at a time allows for small incremental changes, allowing exploration of subtle variations of the original caption. In contrast, applying multiple alternatives simultaneously can produce larger transformations, introducing more diverse scenarios. This approach provides fine-grained control over the extent of modifications, enabling tailored exploration of different levels of change in the generated test cases.

This phase is critically important because it ensures that the generated caption remains coherent and meaningful despite the modifications. Although replacing certain words (such as ``gray'' with ``red'') might seem straightforward, many cases are more complex, requiring careful handling to avoid breaking the sentence's meaning or introducing contradictions. For example, consider a caption like ``a road in a tundra covered in snow during a snowy day''. Replacement of the word ``tundra'' with ``desert'' would result in ``a road in a desert covered in snow during a snowy day'', which is contextually unlikely.

In this step, the LLM is prompted with the following input:

\begin{tcolorbox}[arc=.3em,left=.3em,right=.3em,top=.3em,bottom=.3em]
\begin{center}
\begin{minipage}[t]{.99\linewidth}
\textbf{Prompt}: \textit{
Given the task $<$TASK$>$, modify the caption $<$CAPTION$>$ by applying some of the following transformation described by $<$ALTERNATIVES$>$.
}
\end{minipage}
\end{center}
\end{tcolorbox}

By asking the LLM to generate the new caption directly, rather than applying simple replacement rules from the alternative dictionary, \approach ensures that the LLM processes not only the specific word replacements but also the broader sentence context, maintaining the overall meaning while making necessary adjustments to prevent contradictions or illogical outcomes.
Additionally, by explicitly including the task description at every step of the interaction, the LLM is continuously reminded of the objective it is trying to achieve. This ensures that the generated captions respect the metamorphic relationships inherent in the test case, preserving the critical connections between elements of the image and their semantic meaning.

\begin{figure*}
\centering
    \centering
    \includegraphics[width=.98\textwidth]{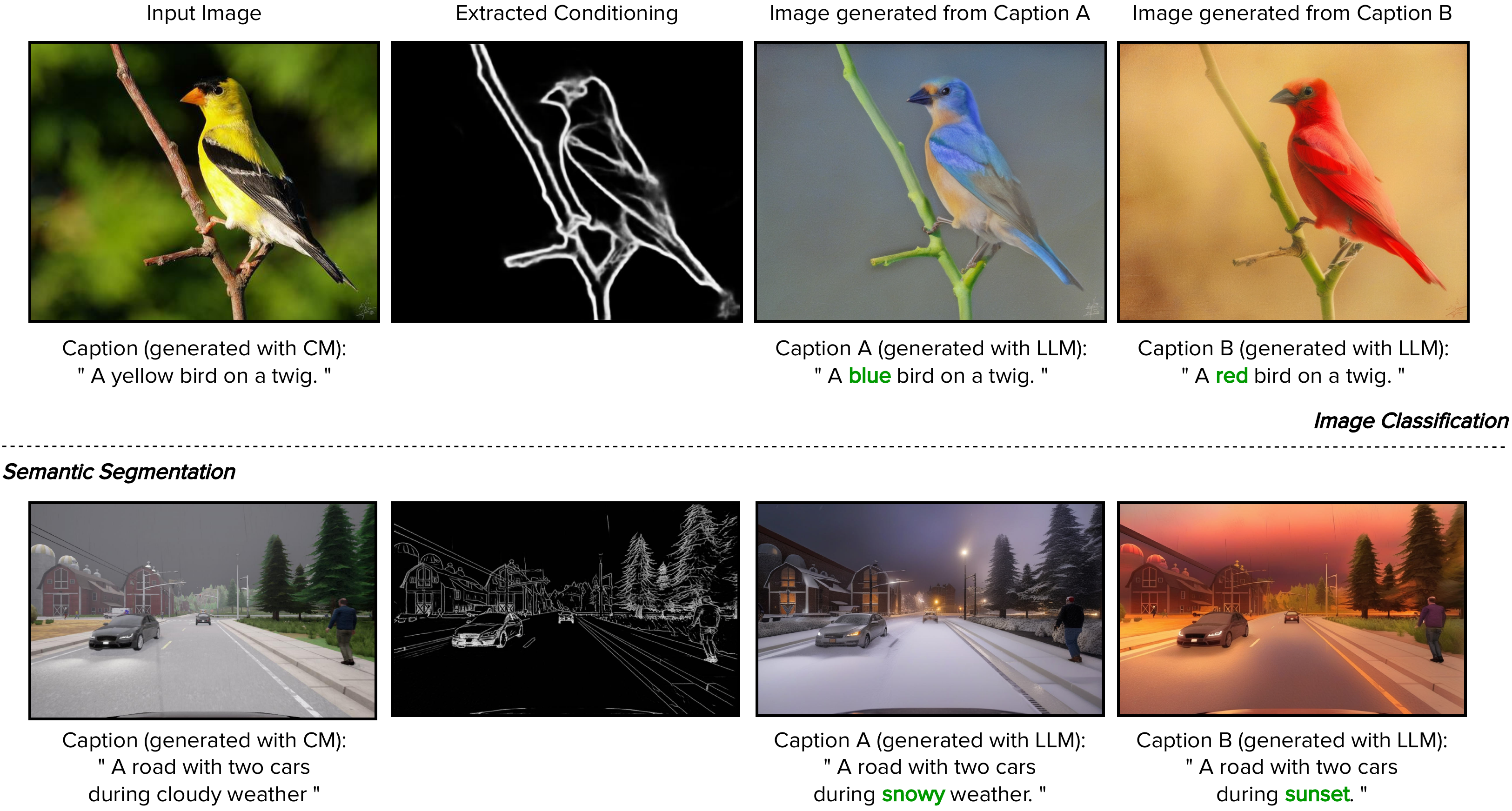}
\caption{Image generation in \approach.}
\label{fig:controlled_generation}
\end{figure*}
\subsection{Controlled Text-to-Image Generation}

The final step of \approach generates a modified image based on the counterfactual caption produced in the previous phase. This step is where the transformation of the image occurs, and it is carried out using a control-conditioned text-to-image diffusion model~\cite{DBLP:conf/iccv/ZhangRA23}.  The key challenge here is not only to generate a new, realistic image that aligns with the counterfactual caption but also to ensure that the spatial structure of the original image is preserved so that the integrity of metamorphic relationships is maintained.

When generating a new test image, the spatial arrangement of key objects and elements must be preserved. For example, in the context of semantic segmentation for autonomous driving, if an image depicts a car driving down a road, the generated image must include the car in the same location as the original image relative to the road, even if its color or weather conditions are changed. This way, the transformations to be applied will only affect specific attributes (e.g., altering weather or object properties) without impacting the fundamental geometry or layout of the scene. On the other hand, a distorted spatial structure could mislead the test results, making it unclear whether a failure is due to the actual shortcomings of the model or due to irrelevant transformations in the image.

To achieve spatial structure preservation, \approach uses control-conditioned diffusion models. These models allow fine-grained control over the generated image by incorporating conditioning inputs that preserve the spatial layout of the original image while applying the desired modifications.

\autoref{fig:controlled_generation} showcases examples of test cases generated by \approach for image classification (top row) and semantic segmentation (bottom row).
For image classification, the input image belongs to the class \textit{bird}, described by the captioning model as ``A yellow bird on a twig''. The second column displays the conditioning input extracted from the original image to preserve spatial arrangements. The remaining columns show images generated from alternative captions produced by the LLM: Caption A (``A blue bird on a twig'') changes the bird color to blue, while Caption B (``A red bird on a twig'') changes it to red. 
These augmentations demonstrate \approach ability to alter specific attributes while maintaining spatial structure and preserving the relevance of the class \textit{bird}. 

For semantic segmentation, the input image depicts a road with two cars during cloudy weather, with the ground truth represented as a semantic map of pixel-level classifications. The captioning model describes it as ``A road with two cars in cloudy weather''. The second column provides the conditioning input to ensure spatial consistency during generation. Caption A (``A road with two cars during snowy weather'') introduces snow to the scene, while Caption B (``A road with two cars during sunset'') applies sunset lighting. Both augmentations preserve the layout of roads, vehicles, and pedestrians as defined by the ground truth semantic map.
\section{Evaluation}
\label{sec:eval}

In this section, we evaluate the performance of \approach and aim to answer the following research questions (RQs):

\noindent\textbf{RQ\textsubscript{1} (Validity).} Can DILLEMA generate valid and realistic test cases from existing data?

\noindent\textbf{RQ\textsubscript{2} (Testing Effectiveness).} Can the generated test cases identify weaknesses in state-of-the-art DL models?

\noindent\textbf{RQ\textsubscript{3} (Retraining).} Can the generated test cases be used to improve the robustness of the tested models?

\subsection{Experimental Setup}
\label{sec:eval:setup}
\noindent\textbf{Datasets.} We performed experiments using two datasets: ImageNet1K~\cite{DBLP:conf/cvpr/DengDSLL009} and SHIFT~\cite{DBLP:conf/cvpr/SunSPWGSTY22}. These datasets represent two different tasks, image classification, and semantic segmentation, allowing us to assess the flexibility and applicability of \approach in various scenarios. ImageNet1K is a large-scale dataset commonly used for image classification tasks and SHIFT is a synthetic dataset designed for evaluating autonomous driving systems under different conditions (e.g., weather changes, lighting conditions).

\noindent\textbf{Tested Models.} We used \approach to test several DL architectures.
For ImageNet1K, we evaluated classification models (that is, ResNet18, ResNet50, and ResNet152~\cite{DBLP:conf/cvpr/HeZRS16}) using pre-trained versions provided by PyTorch. For SHIFT, we tested a semantic segmentation model (i.e., DeepLabV3~\cite{DBLP:conf/eccv/ChenZPSA18} model with a ResNet50 backbone), which we custom-trained following the original authors' training procedure~\cite{DBLP:conf/eccv/ChenZPSA18}. The training of this model took approximately $24$ hours to complete.

\noindent\textbf{Evaluation Metrics.} We used accuracy to evaluate the quality of classification models (on ImageNet1K), and we used mean Intersection over Union (mIoU) to measure the ability to evaluate the quality of semantic segmentation models.

\noindent\textbf{\approach Configuration\footnote{
To support reproducibility, all our data, including the code of \approach, the results of the human survey, of the testing and retraining, are available in our replication package: \url{https://github.com/deib-polimi/dillema}.}.} We used BLIP2 6.7B~\cite{DBLP:conf/icml/0008LSH23} as the captioning model to generate context-aware descriptions, chosen for its ability to produce detailed, semantically rich captions. As LLM, we selected a 5-bit quantized LLaMA-2 13B~\cite{DBLP:journals/corr/abs-2307-09288} model to identify keywords, generate alternatives, and create counterfactual captions. We chose LLaMA-2 because it is open source and effective, and we opted for the 13B version with 5-bit quantization since it provided a balance between performance and resource efficiency given our computational and cost constraints.
Lastly, for image generation, we used ControlNet~\cite{DBLP:conf/iccv/ZhangRA23} with edge conditioning, a control-conditioned text-to-image diffusion model. ControlNet enabled us to introduce modifications to the images while maintaining the spatial structure of the original scene, ensuring that the relationships between objects and their surroundings remained consistent.
Although we chose these general-purpose models for compatibility with consumer hardware and reasonable runtime, other models with different capabilities could be used depending on specific needs.

\noindent\textbf{Prompt Template.} To guide the LLM effectively, we used a one-shot in-context learning approach~\cite{DBLP:conf/nips/Wei0SBIXCLZ22}, where each prompt included an example to help the model understand the request more accurately. The example illustrated the expected input and output formats. Each prompt was constructed to provide context and explicitly instruct the LLM on the required output format, which allowed for automated post-processing. If the LLM response failed to adhere to the specified output format and could not be automatically parsed, we repeated the request with a different random seed. This iterative process continued until a parsable response was obtained.

\noindent\textbf{Retraining Settings.}
For ImageNet1K, we re-trained the ResNet models using a batch size of $100$ and the SGD optimizer with an initial learning rate of $0.1$, a momentum of $0.9$ and a weight decay of $1 \times 10^{-4}$. The learning rate was decayed using the PyTorch StepLR scheduler with a step size of $30$ and a gamma of $0.1$, over $90$ epochs.
For SHIFT, we re-trained the DeepLabV3 model using the original settings provided by its authors. Specifically, the batch size was set to $12$, with training conducted over $200$ epochs using the Adam optimizer with a learning rate of $0.002$, betas set to $(0.9, 0.999)$, and epsilon set to $1 \times 10^{-8}$.

\noindent\textbf{Hardware and Software.} The experiments were carried out on an AWS virtual machine with an A10G NVIDIA GPU with 24GB of memory. Neural networks were designed using PyTorch 2.0.1, and accelerated using CUDA 11.8.
In general, the empirical evaluation required about $120$ GPU hours.
$96$ GPU-hours were spent on Imagenet1K ($125,000$ test cases), $24$ GPU hours were spent on SHIFT ($10,000$ test cases).

\subsection{RQ\textsubscript{1}. Validity}
\label{sec:experiments:validity}
This experiment aims to evaluate the realism and validity of the generated images, ensuring that they preserve the metamorphic relationship for both datasets and assessing how often hallucinations occur due to potential errors during the five steps of \approach. By validating the generated images end-to-end, we aim to identify instances where the pipeline produces incorrect or unrealistic results.
To achieve this, we conducted a human study using Amazon Mechanical Turk. Human evaluators were asked to verify if the generated images preserved the metamorphic relationship for both datasets.

In total, we obtained $2,500$ total responses. To ensure quality, we used control questions to filter unreliable answers. Responses failing these quality checks were discarded. To ensure experienced participants, the workers were selected based on an approval rate greater than $95\%$ and at least $50$ completed tasks. Each image was evaluated by five independent workers and the questions were discarded if consensus (agreement of at least $\frac{4}{5}$ participants) was not reached. In the end, only $2,380$ responses (out of $2,500$) were considered robust and good enough to answer the research question.

\begin{figure}[h]
    \centering
    \includegraphics[width=.45\textwidth]{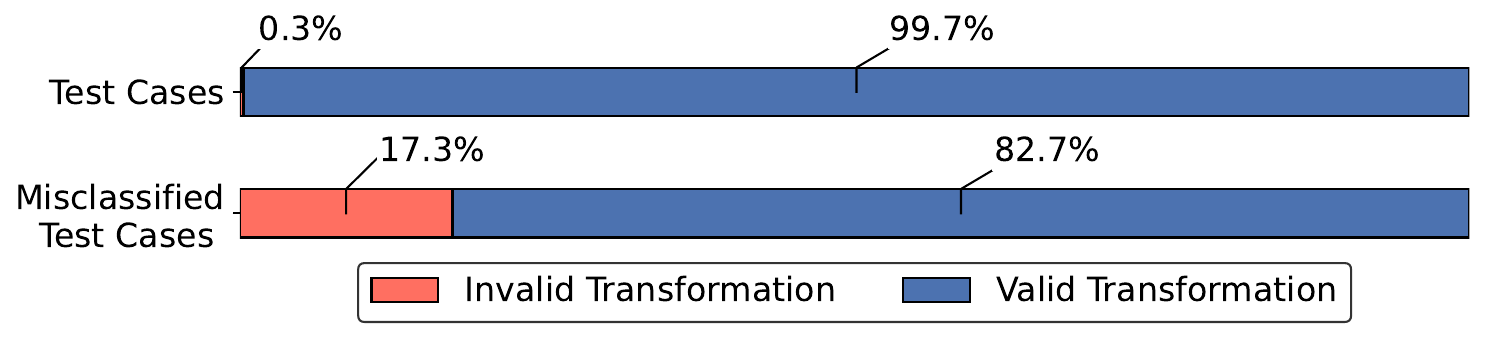}
    \caption{Validity of the Generated Test Cases for Classification.}
    \label{fig:rq1_validity_imagenet1k}
\end{figure}

For ImageNet1K (\autoref{fig:rq1_validity_imagenet1k}), we used two types of questions and considered a transformation to be valid if our approach were able to correctly augment an existing image without modifying the label associated with it.
First, we performed a general evaluation on a randomly sampled set of $300$ augmented images from all generated cases to measure the overall validity.
Then, we proposed a focused evaluation of $100$ augmented images that the ResNet18 model misclassified, to check if the images were valid and interpretable by humans even when misclassified by the model.

Our human study shows that human assessors achieved agreement on all images and $99.7\%$ of the augmented images were correctly classified by human assessors. Of the $300$ images, only $1$ image did not preserve the label associated with the original image.
For the set of images where the model (i.e., ResNet18) produced a misclassification, $82.7\%$ were still considered valid by human evaluators. This shows that while the test cases generated by \approach effectively induced misclassifications in the model, most of them could still be correctly classified by humans. This suggests that failures can often be attributed to bugs in the model rather than flaws in the image generation process, reinforcing the validity and utility of \approach for robust model testing.

\begin{figure}[h]
    \centering
    \includegraphics[width=.45\textwidth]{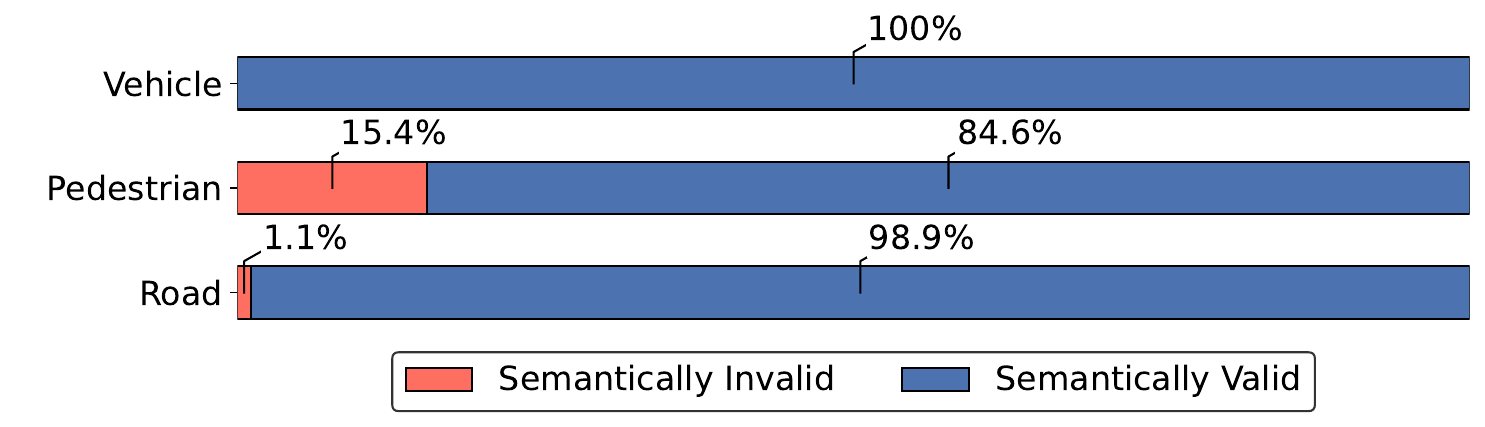}
    \caption{Validity of the Generated Test Cases for Driving.}
    \label{fig:rq1_validity_shift}
\end{figure}

For the SHIFT dataset (\autoref{fig:rq1_validity_shift}), we randomly selected $100$ augmented images. Among these, all depicted roads, $25$ included vehicles, and $15$ featured one or more pedestrians. Evaluators were tasked with verifying whether key elements critical for autonomous driving, such as roads, vehicles, and pedestrians, were consistently preserved through the transformations.  We checked these aspects since they are key elements that influence the behavior of an autonomous driving system.

We observed the following validity rates: road preservation at $98.9\%$ ($100$ questions, $7$ were discarded due to lack of consensus), pedestrian preservation at $84.6\%$ ($15$ questions, $2$ discarded due to lack of consensus), and vehicle preservation at $100.0\%$ ($25$ questions, $1$ discarded due to lack of consensus). These results highlight that \approach can effectively maintain certain features, such as roads and vehicles, while being slightly less effective at preserving pedestrians.

\subsection{RQ\textsubscript{2}. Testing Effectiveness}

\begin{figure*}
\centering
    \begin{subfigure}[b]{.49\textwidth}
    \centering
\includegraphics[width=\textwidth]{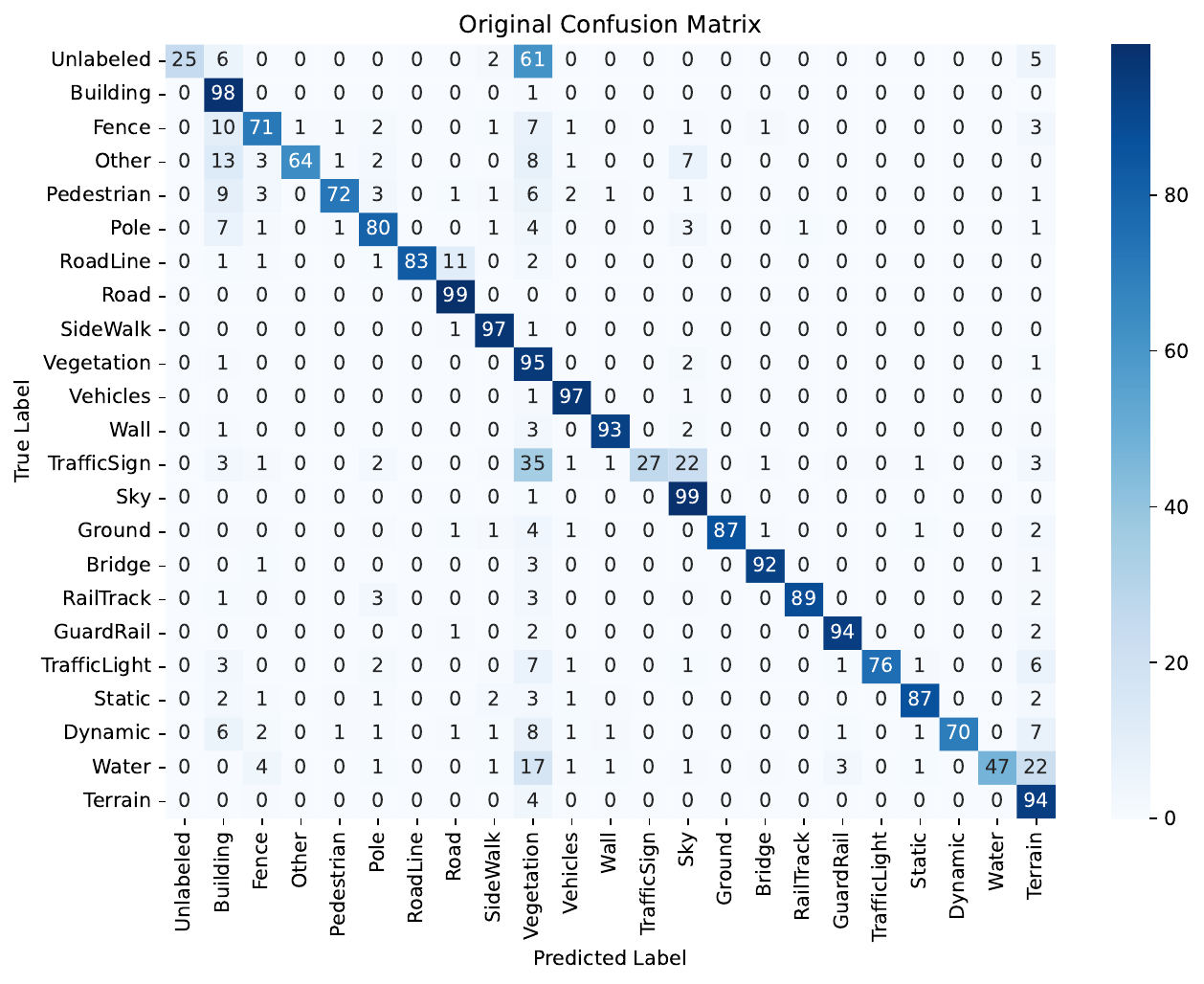}
    \caption{Accuracy on Original Test Suite.\label{fig:image1}}
    \end{subfigure}
    \begin{subfigure}[b]{.49\textwidth}
    \centering
    \includegraphics[width=\textwidth]{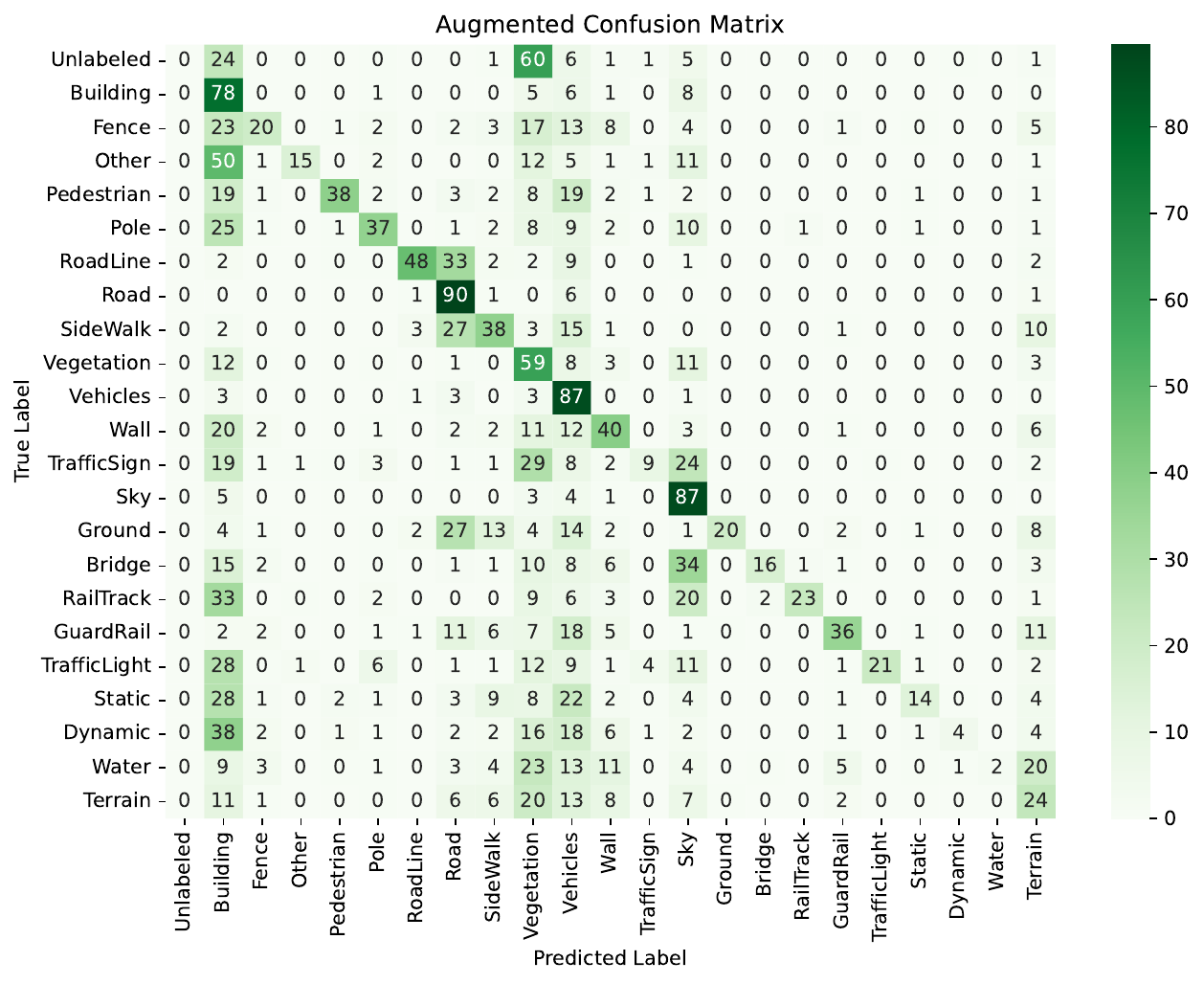}
    \caption{Accuracy on \approach Augmented Test Suite.\label{fig:image2}}
    \end{subfigure}
\caption{Multi-class Confusion Matrix.}
\label{fig:rq2_effectiveness}
\end{figure*}

To evaluate the effectiveness of \approach, we evaluated its ability to detect weaknesses in state-of-the-art DL models using the generated test cases.

First, we performed experiments on ImageNet1K, focusing on identifying misclassification errors. For this purpose, we augmented $25$ images for each of the $1,000$ classes in the dataset. Each image was augmented five times to take advantage of the stochastic nature of diffusion models, which can generate different augmentations from the same input. The performance of the test suite generated by \approach was compared with the test set already available in the dataset.

\begin{table}[h]
    \footnotesize
    \centering
    \begin{tabular}{@{}lcc@{}}
        \toprule
        \xspace\space\xspace\space  \textbf{Architecture}\xspace\space\xspace\space               & \xspace\space\xspace\space\xspace\space\xspace\space\textbf{Original Test Suite}\xspace\space\xspace\space\xspace\space\xspace\space                      & \xspace\space\xspace\space\textbf{\approach Test Suite}\xspace\space\xspace\space \\ \midrule
        \xspace\space\xspace\space  ResNet18\xspace\space\xspace\space & 5.26\% & 53.29\% \\
        \xspace\space\xspace\space  ResNet50\xspace\space\xspace\space & 2.55\% & 45.47\% \\
        \xspace\space\xspace\space  ResNet152\xspace\space\xspace\space & 1.47\% & 42.33\% \\ \bottomrule
        \end{tabular}
    \caption{Test Effectiveness.}
    \label{tab:rq2_effectiveness}
\end{table}

\autoref{tab:rq2_effectiveness} reports the performance of three ResNet variants in both test suites. The results reveal that, on average, $3.1\%$ of the original test suite was able to highlight misbehaviors, while $47.0\%$
of the test suite generated by \approach exposed faulty behaviors. However, it is important to note that, as discussed in \autoref{sec:experiments:validity}, not all of these detected misbehaviors may represent true failures. The human study confirmed that approximately $82.7\%$ of the misbehaviors detected by \approach were valid failures. Even after normalizing for this factor, the effectiveness of \approach remains significantly higher ($38.9\%$) than the original test set.

In addition, we analyzed how many augmentations per image led to model errors. Our findings indicate that for $33.29\%$ of the images, all augmentations resulted in misclassifications, whereas for $24.85\%$, none of the augmentations caused errors.

For the SHIFT dataset, we evaluated the DeepLabV3 model on the semantic segmentation task. The evaluation compared the augmented test set created by \approach with the original SHIFT test set. \autoref{fig:rq2_effectiveness} presents the normalized multi-class confusion matrix of the tested model on the original and augmented data. Rows represent the ground truth, columns represent the predicted class, and the diagonal indicates the percentage of correct predictions.

The results show that \approach successfully exposed interesting faulty behaviors. For example, in semantic classes where the model appeared robust in the original dataset, such as \textit{SideWalk} ($97\%$ correctly classified), the model showed significant vulnerability in the augmented dataset (only $38\%$). In more critical classes such as \textit{Road} and \textit{Vehicle}, we observed that the model maintained a relatively robust performance, with errors increasing by $9\%$ and $10\%$, respectively, as the accuracy decreased from $99\%$ and $97\%$ in the original dataset to $90\%$ and $87\%$ in the augmented dataset. However, for pedestrian recognition, the augmented dataset revealed a much higher vulnerability, with $34\%$ more misclassifications compared to the original dataset. This highlights the need to retrain the model with a stronger focus on identifying pedestrians to address this critical weakness.

These results highlight that \approach not only highlights hidden vulnerabilities in classes previously considered robust but also provides insights into critical performance degradations in safety-relevant semantic classes. In general, \approach effectively exposes model weaknesses in various scenarios.

\subsection{RQ\textsubscript{3}. Retraining Robustness}

To assess whether the test cases generated by \approach can improve the robustness, we conducted retraining experiments using the synthetically generated data. Retraining aimed to evaluate whether the incorporation of augmented test cases into the training process leads to improved performance on both original and augmented data.

For the ImageNet1K dataset, we retrained the ResNet18 model using a combined training set consisting of the original data and the augmented test cases generated by \approach. The model was re-trained for 90 epochs using the settings described in \textit{Retraining Settings}. The re-trained model showed a significant improvement in robustness, achieving a $52.27\%$ increase in accuracy in the augmented test cases and a $20.19\%$ improvement in the original test suite.

Concerning SHIFT, we achieved an improvement in mIoU across the original and augmented test sets. After retraining, mIoU in the original test suite improved from $85.32\%$ to $88.76\%$, while mIoU in the augmented dataset showed a more pronounced increase from $72.45\%$ to $80.32\%$.
Specifically, the retraining process revealed that while performance degradation on critical semantic classes like \textit{Road} and \textit{Vehicles} was minor, pedestrian recognition showed a significant recovery, increasing from $38\%$ to $62\%$. This improvement highlights the value of \approach in augmenting datasets to address vulnerabilities in safety-critical tasks.

These findings demonstrate that the generated test cases are highly effective in not only uncovering model vulnerabilities but also improving the robustness of DL models when incorporated into the retraining process.

\subsection{Threats to Validity}
\noindent\textbf{Internal Validity.}
Our pipeline relies on pre-trained models (captioning, LLM, diffusion) and random sampling of alternatives, which can introduce randomness and potential skew (e.g., consistently generating “red” vehicles). Another concern is the domain shift between real images and our synthesized outputs: models might perform worse simply because of unfamiliar synthetic characteristics rather than true weaknesses. However, our human study indicates that the vast majority of generated images retain labels recognizable to human evaluators, suggesting that they are semantically coherent rather than purely artificial or misleading. Thus, while some failures could stem from synthetic artifacts, the high human agreement on these images implies that many observed misclassifications reflect genuine model vulnerabilities rather than artifacts alone.

\noindent\textbf{External Validity.}
We tested \approach on classification and segmentation from distinct domains, but it may not generalize to specialized scenarios (e.g., medical imaging). Although each component (captioning, LLM, diffusion) seems broadly applicable, further testing on diverse datasets is required to confirm adaptability for industrial use and other vision tasks.

\noindent\textbf{Construct Validity.}
Our primary measure of success is whether the generated images preserve ground-truth labels and uncover vulnerabilities. While human assessments indicate that images remain valid, potential biases in LLM-generated alternatives (e.g., color choices) could distort conclusions. Additionally, the notion of validity is subjective; thus, future work should employ more rigorous metrics or automated checks to validate semantic consistency in generated test cases.
\section{Related Work}
\label{sec:related}

Metamorphic testing has emerged as an effective approach to test DL-based systems without explicit test oracles~\cite{DBLP:conf/icse/ZhouLKGZ0Z020, DBLP:journals/pacmse/ChenJYGZC24}.
Notably, DeepTest~\cite{DBLP:conf/icse/TianPJR18} applies a simple image transformations, such as brightness and contrast adjustments, translations, rotations, and blurs, to generate synthetic images that represent real-world conditions. DeepRoad~\cite{DBLP:conf/kbse/ZhangZZ0K18} is a metamorphic testing approach that generates various driving conditions (foggy and snowy) using complex computer vision techniques such as GAN.
DeepXplore~\cite{DBLP:journals/cacm/PeiCYJ19} employs a white-box testing approach, with the aim of increasing neuron coverage and uncovering inconsistent behaviors in different models under the same input conditions. 
Existing approaches either rely on simple transformations that may not replicate real-world effects or use complex methods like GANs, which require extensive training and scenario-specific data collection. In contrast, \approach generates diverse and realistic test cases without the need for ad-hoc training, significantly improving the scope and applicability of metamorphic testing.

Data augmentation~\cite{DBLP:journals/corr/abs-2407-04103} is commonly used during training to improve model robustness by generating various variations of existing data. Recent advances in language-guided and diffusion-based methods have enabled sophisticated augmentations, often preserving spatial structure and semantic consistency~\cite{DBLP:journals/corr/abs-2409-13661,DBLP:conf/icst/LambertenghiS24}. For example, Dataset Interfaces~\cite{DBLP:journals/corr/abs-2302-07865} alter minor aspects such as backgrounds to simulate distribution shifts while maintaining class relevance. ALIA~\cite{DBLP:conf/nips/DunlapUZYGD23} combines image captioning and language models to create semantic integrity-targeted augmentations for robust training. While Dataset Interfaces focus on shifting contextual factors and ALIA operates on sets of images to augment training data, \approach takes a more granular approach by captioning each image individually. This allows \approach to explore modifications customized to the specific context of each image.
\section{Conclusion}
\label{sec:conclusion}

In conclusion, the synergy of captioning, LLM-driven counterfactuals, and control-conditioned diffusion effectively reveals model weaknesses and increases robustness. Future work will compare with additional baselines and explore prioritization of the generated test cases.

\bibliographystyle{IEEEtran}  
\balance
\bibliography{bibl}  

\end{document}